\pdfoutput=1
\documentclass[10pt, a4paper]{article}

\usepackage[final]{lrec2026} 

\usepackage{multicol}
\usepackage{multirow}
\usepackage{makecell}
\usepackage{glossaries}
\usepackage{hyperref}
\usepackage{cleveref}

\usepackage{enumitem}
\setlist[itemize,enumerate]{itemsep=1.5pt, topsep=1.5pt, parsep=1.5pt, partopsep=1.5pt}

\makeglossaries
\glsdisablehyper

\newacronym{aed}{AED}{attention encoder decoder}
\newacronym{asr}{ASR}{automatic speech recognition}
\newacronym{bpe}{BPE}{byte-pair encoding}
\newacronym{ctc}{CTC}{connectionist temporal classification}
\newacronym{fh}{FH}{factored hybrid}
\newacronym{g2p}{G2P}{grapheme-to-phoneme}
\newacronym{gmm}{GMM}{Gaussian-mixture-model}
\newacronym{lm}{LM}{language model}
\newacronym{oov}{OOV}{out-of-vocabulary}
\newacronym{rnnt}{RNN-T}{recurrent neural network transducer}

\newcommand\T{\rule{0pt}{1.80ex}}

\title{Supplementary Resources and Analysis for Automatic Speech Recognition Systems Trained on the Loquacious Dataset}

\name{Nick Rossenbach$^{\ast}$$^{\dagger}$, Robin Schmitt$^{\ast}$$^{\dagger}$, Tina Raissi$^{\ast}$, \\ {\bf \large Simon Berger$^{\ast}$$^{\dagger}$, Larissa Kleppel$^{\ast}$, Ralf Schlüter$^{\ast}$$^{\dagger}$}} 

\address{$^{\ast}$RWTH Aachen University, $^{\dagger}$AppTek.ai\\
         Aachen, Germany \\
         \{\textit{lastname}\}@ml.rwth-aachen.de}

\abstract{
The recently published Loquacious dataset aims to be a replacement for established English \gls{asr} datasets such as LibriSpeech or TED-Lium.
The main goal of the Loquacious dataset is to provide properly defined training and test partitions across many acoustic and language domains, with an open license suitable for both academia and industry.
To further promote the benchmarking and usability of this new dataset, we present additional resources in the form of $n$-gram \glspl{lm}, a \gls{g2p} model and pronunciation lexica, with open and public access.
Utilizing those additional resources we show experimental results across a wide range of \gls{asr} architectures with different label units and topologies.
Our initial experimental results indicate that the Loquacious dataset offers a valuable study case for a variety of common challenges in \gls{asr}.
\\ \newline  }

\begin{document}

\maketitleabstract

\section{Introduction}

In order to compare \gls{asr} systems across scientific publications, it is necessary to clearly define the training and testing conditions and data used. 
For this purpose, many benchmarks and corresponding datasets have been published in the past.
Notable datasets for academic use covering English language are: Switchboard \cite{switchboard}, LibriSpeech \cite{librispeech} or TED-Lium \cite{DBLP:conf/specom/HernandezNGTE18}. 
Especially LibriSpeech, with more than 1700 citations in 2024 alone\footnote{according to Google Scholar, 15th Oct. 2025}, is dominating \gls{asr} research.
However, due to the increase in computation power, more and more research is performed without any constraints on the used data.
This generally yields better results, but limits scientific comparability. 
On the other hand, existing datasets only provide a smaller amount of data, 
and are often not realistic in the diversity of audio and text conditions.
Also, some datasets are not directly accessible, were removed from public access after publication\footnote{The TED terms of use disallow machine learning usage, existing datasets based on TED were taken offline}, or have restricted licenses.

The Loquacious dataset \cite{parcollet25_interspeech} was published with the idea in mind to replace older datasets, offering multiple benefits. The important highlights are:

\begin{itemize} [leftmargin=2em]
	\item A diverse selection of data sources containing both read and spontaneous speech
	\item Three different training subsets, consisting of 250, 2.5k, and 25k hours, with 250 and 2.5k having uniformly balanced data sources
	\item Defined dev and test sets for every data source
	\item A strong and consistent text normalization
	\item Open licensing for academia and industry
\end{itemize}
Such aspects make the Loquacious dataset a promising replacement for older standard datasets, with the 2.5k hours subset having a similar magnitude as the 960 hours of LibriSpeech.
Moreover, due to similar text normalization, it is straightforward to use the dataset with existing LibriSpeech-oriented pipelines. 

As the original release of Loquacious only included training and test data, we aim to extend the available material with count-based \glspl{lm}, pronunciation lexica, and pre-trained \gls{g2p} models.
Such resources exist for LibriSpeech\footnote{\url{https://openslr.org/11/}} and other datasets,
and are relevant for the comparison of \gls{asr} pipelines that require such resources.
The introduction of a pronunciation lexicon allows for explicit decoder representations via lexical trees or finite state machines.
These representations can integrate word-level and lexical information together with model specific label topologies~\cite{mohri2002weighted}.
This type of constraint could benefit the \gls{asr} performance, for example under domain mismatch conditions~\cite{raissi2025analysis}.
Moreover, the alignment of models using a blank-free label topology with phonemic units and silence modeling offers the advantage of providing word and phoneme boundaries unambiguously.
Prior works also show that phoneme-based models are able to provide alignments with more accurate time information relative to the evidence in the signal, compared to the \gls{bpe} based models~\cite{rousso2024tradition,jiang2023neural}.


\subsection{Contributions}

\begin{table*}[t]
	\begin{center}
		\setlength{\tabcolsep}{0.29em}\renewcommand{\arraystretch}{1.1} 
		\caption{Perplexities and \gls{oov} percentage of the different count-based \glspl{lm} on the respective dev sets. All \glspl{lm} are restricted to the 216k words vocabulary.}
		\vspace{0.5em}
		\label{tab:lmppl}
		\begin{tabular}{|c|c||c|c||c|c|c|c|c|c|c|c|}
			\hline
			\multirow{2}{*}{Language Model} &
			\multirow{2}{*}{\makecell{$n$-gram\\count}} & \multicolumn{2}{c||}{Loquacious} & \multicolumn{2}{c|}{Commonvoice} & \multicolumn{2}{c|}{LibriSpeech} & \multicolumn{2}{c|}{VoxPopuli} & \multicolumn{2}{c|}{Yodas}\T\\
			\cline{3-12}
			& & $\text{OOV}_{[\%]}$ & PPL  & $\text{OOV}_{[\%]}$ & PPL  & $\text{OOV}_{[\%]}$ & PPL  & $\text{OOV}_{[\%]}$ & PPL  & $\text{OOV}_{[\%]}$ & PPL\T\\
			\hline 
			\hline
			3-gram pruned & 36M & 0.58 & 222 & 1.25 & 361 &0.58 &231 & 0.13 & 154 &0.24 &201\T\\
			\hline
			4-gram pruned & 54M & 0.58 & 202 & 1.25 &327 & 0.58 &211 & 0.13 &139 & 0.24 &187\T\\
			\hline
			+ LibriSpeech & 201M & 0.58 & 196 & 1.27 & 352 & 0.58 & 149 & 0.13 & 173 & 0.24 & 243\T\\
			\hline
			4-gram unpruned & 246M & 0.58 & 193 & 1.25 & 311 & 0.58 & 195 & 0.13 &135 &0.24 &182\T\\
			\hline
		\end{tabular}
		\vspace{-1em}
	\end{center}
\end{table*}

With this work, we benchmark several \gls{asr} architectures with different label topologies using \gls{bpe} and phoneme label units.\ We show the effect of different decoding methods, with open or closed vocabulary and optional use of a \gls{lm}.
For this purpose, we extend the Loquacious dataset with a pronunciation lexicon, a Sequitur \cite{Bisani-2008-Joint-sequencemodel} based \gls{g2p} model and pre-trained $n$-gram count-based \glspl{lm}.
All resources are available as part of the official Loquacious repository on HuggingFace\footnote{\url{https://huggingface.co/datasets/speechbrain/LoquaciousSet}}.
The scripts for creating the resources are available on Github\footnote{\url{https://github.com/rwth-i6/LoquaciousAdditionalResources}}.

We provide results using lexicon constrained and 4-gram LM decoding across many different ASR architectures for the 250 hours and 2.5k hours subsets of Loquacious.
In addition, we include additional ablation studies regarding the pronunciation lexicon and the effect of pronunciation variants.
As the original publication for Loquacious mostly contained results without data augmentation, we investigate the effect of SpecAugment \cite{DBLP:journals/corr/abs-1904-08779} and speed perturbation \cite{ko15_interspeech}.
Finally, we discuss additional insights we gained about this new dataset while working with it.

%

\section{Language Modeling and Lexicon}

Despite the prominence of neural (large) language models in \gls{asr}, count-based \glspl{lm} still have their advantages as they are fast to train and need often only negligible computational resources.
Just recently NVidia explored GPU support for count-based \glspl{lm} \cite{bataev25_interspeech}.
 Count-based \glspl{lm} are suitable for resource efficient analysis of how well training data is matching the test data.

\subsection{Vocabulary}
\label{sec:vocab}

Word-level \glspl{lm} require a fixed vocabulary. We aim for a size that is similar to that of the official lexicon for LibriSpeech, which includes 200k words.
In order to avoid an arbitrary cutoff, we choose the following process to determine the vocabulary:
\begin{enumerate}[leftmargin=2em]
	\item We use the CMU Pronunciation Dictionary (CMUdict)\footnote{\url{http://www.speech.cs.cmu.edu/cgi-bin/cmudict}} 0.7b as initial starting point for our vocab, which contains pronunciations for about 124k words, with 134k total pronunciations.
	\item We select every word from the CMUdict that is appearing at least once in the full text of the Loqacious \textit{train-large} training data.
	This results in 95k words.
	\item We then select every word that appears at least 4 times in the \textit{train-large} training data, which gives us 197k words.
	\item The union of both results in 216k words which we use as target vocabulary.
\end{enumerate}

\subsection{Pronunciation Lexicon}
\label{sec:lex}

Characters or sub-word methods such as \gls{bpe} \cite{sennrich-etal-2016-neural} or Sentencepieces via unigram \gls{lm} \cite{kudo-2018-subword} nowadays are common for target label representations in \gls{asr}.
To additionally allow for phoneme representations, we create phoneme pronunciations for all words in the vocabulary file as follows:
\begin{enumerate}[leftmargin=1em]
	\item For all words that are part of the CMUdict, we take the existing entries (with one or more pronunciations) and remove stress markers.
	\item We train a 5th-order \gls{g2p} model with Sequitur using the full CMUdict as training data.
	For higher orders, we see no further improvement on our held-out cross-validation set.
	\item We generate two pronunciation variants using Sequitur for all words which are not in the CMUdict.
\end{enumerate}
For the last step, we either take only the most probable variant,
or we select the second variant based on a probability threshold which is set to either 60\% or 80\%. We only consider a second pronunciation variant generated if the re-normalized posterior probability of the first variant is below this threshold.

\subsection{Count-based Language Models}
\label{sec:countlm}

We use the KenLM toolkit \cite{heafield-2011-kenlm} to train ARPA-format count-based language models.
We create \glspl{lm} in three different settings:
\begin{enumerate}[leftmargin=2em]
	\item 3rd-order with singleton pruning of 3-grams.
	\item 4th-order with singleton pruning of 3-grams and 4-grams.
	\item 4th-order without any pruning.
\end{enumerate}
The perplexities and \gls{oov} rates can be found in \Cref{tab:lmppl}.
As expected, every test-set has different perplexities, indicating both the complexity of the text and how much fitting training data is available within the full Loquacious dataset.


\section{Automatic Speech Recognition}

The original Loquacious publication only listed results for \gls{asr} systems that combine \gls{aed} \cite{chan2016las} and  \gls{ctc} \cite{10.1145/1143844.1143891}. 
We use a set of different systems with different training pipelines to verify our lexicon and $n$-gram \glspl{lm}, as well as showing open-vocabulary results and other ablation studies.
\gls{asr} systems using \gls{ctc} loss only have an encoder with independent label outputs.
Such systems tend to profit the most from additional \glspl{lm} as they have no explicit language context modeling.
For comparison, we include \gls{rnnt}-based \cite{DBLP:journals/corr/abs-1211-3711} \gls{asr} systems.
They provide a decoder structure for explicit language context modeling, so that additional language model information might be less relevant.
To include a setup which properly supports pronunciation variants, we also include a \gls{fh} model~\cite{raissi2020fh}. \gls{fh} is a modern alternative to the classic hybrid neural network hidden Markov model~(HMM)~\cite{Bourlard+Morgan:1993}, being trained from scratch and thus avoiding both initial alignments from Gaussian mixture HMMs and decision tree based phone clustering.
The diphone \gls{fh} is trained similarly to CTC but offers a blank-free label topology with phonemic units and diphone context modeling.
Finally, we include an \gls{aed} system to be able compare to the original Loquacious publication, even if in our case it does not support word level count-\gls{lm} decoding.


%

%

\subsection{General Settings}

	\begin{table*}[t]
	\begin{center}
		\caption{Recognition results for \gls{bpe} ASR systems that work without additional \glspl{lm} or lexicon. Abbreviations for test sets: Commonvoice (CV), LibriSpeech (LS), VoxPopuli (VP), Yodas (YD).}
		\vspace{0.5em}
		\label{tab:baseline}
		\begin{tabular}{|c|c|c||c|c||c|c|c|c|}
			\hline
             \multirow{3}{*}{\makecell{Data}} & \multirow{3}{*}{\makecell{Architecture}} &  \multirow{3}{*}{\makecell{Decoding}} & \multicolumn{6}{c|}{WER [\%]}\T\\ \cline{4-9}
            & & & \multicolumn{2}{c||}{Loquacious} & CV & LS & VP & YD\T\\ \cline{4-9}
            & & & dev & test  & \multicolumn{4}{c|}{test} \T\\
			\hline 
			\hline
			\multirow{5}{*}{\makecell{\textit{train.small} \\ 250h}} & CTC & \multirow{2}{*}{Greedy} &  17.2 & 18.6 & 30.1 & 16.7 & 12.5 & 21.1\T\\
			\cline{2-2}
			\cline{4-9}
			& \multirow{2}{*}{RNN-T} & &  15.2 & 16.1 & 26.8 & 14.1 & 11.0 &17.9\T\\
			\cline{3-9}
			& & Beam-Search & 14.9 & 15.9 & 26.3 & 13.9 & 10.9 & 17.6\T\\
			\cline{2-9}
			& \multirow{2}{*}{AED} & Greedy & 15.3 & 16.7 & 28.0 & 14.7 & 10.8 & 20.0\T\\
			\cline{3-9}
			& & Beam-Search & 14.7 & 16.0 & 26.8 & 14.2 & 10.4 & 18.0\T\\
			\hline
			\hline
			\multirow{5}{*}{\makecell{\textit{train.medium} \\ 2500h}} & CTC & \multirow{2}{*}{Greedy} & 10.6 & 11.5 & 19.3 & \phantom{0}9.0 & \phantom{0}8.1 & 15.4\T\\
			\cline{2-2}
			\cline{4-9}
			& \multirow{2}{*}{RNN-T} & & \phantom{0}9.0	& 10.3 & 16.1 & \phantom{0}7.2 & \phantom{0}7.0 & 21.9\T\\
			\cline{3-9}
			& & Beam-Search & \phantom{0}8.4 & \phantom{0}9.3 & 15.4 & \phantom{0}7.0 & \phantom{0}6.9 & 12.4\T\\
			\cline{2-9}
			& \multirow{2}{*}{AED} & Greedy & \phantom{0}8.7 & 10.0 & 16.5 & \phantom{0}7.5 & \phantom{0}7.0 & 15.1\T\\
			\cline{3-9}
			& & Beam-Search & \phantom{0}8.4 & \phantom{0}9.5 & 15.9 & \phantom{0}7.3 & \phantom{0}7.0 & 13.1\T\\
			\hline
		\end{tabular}
		\vspace{-1em}
	\end{center}
\end{table*}

The \gls{asr} systems in this work are based on different software and settings, covering a broader range of different pipelines.
All neural models are trained using RETURNN \cite{zeyer2018:returnn} or variants of it\footnote{\url{https://github.com/JackTemaki/MiniReturnn}}. The \gls{fh} system pipeline uses a Tensorflow backend, while all other models are based on PyTorch.
All experiment pipelines are based on Sisyphus \cite{DBLP:conf/emnlp/PeterBN18}.
Prefix-tree based decoding is performed using Flashlight \cite{kahn2022flashlight} via the \textit{torchaudio} interface for \gls{ctc} and RASR \cite{rybach2011rasr} for (m)RNN-T and \gls{fh}.
All systems use a 12-layer Conformer \cite{conformer} encoder with a hidden size of 512.
We use 80-dimensional log-mel spectrograms with a window size of 25ms and a shift of 10ms including SpecAugment and speed perturbation, with exception of \gls{fh} that uses only SpecAugment.
The features are down-sampled with two convolutional layers by a factor of six for \gls{rnnt} and \gls{aed} and four for the other systems.
We train each system for 100 epochs on the 250 hours \textit{train.small} subset and for 40 epochs on the 2.5k hours \textit{train.medium} subset.
The training pipelines include different variants of Adam and one-cycle-learning-rate (OCLR) scheduling.
We evaluate each model on the full dev and test sets, and in many cases report the word error rates (WER) individually for the 4 different test sets.
For any system using a \gls{lm}, we perform tuning of the \gls{lm}  and label prior correction scales on a fixed random subset of the full dev set.
When using phonemes, we use an additional label set for end-of-word phonemes.

\subsection{CTC}

The \gls{ctc} system has a total parameter count of 77M.
For the \gls{bpe}-based systems, three different decoding modes are used:
\begin{enumerate}[leftmargin=2em]
	\item Simple greedy decoding by picking the position-wise maximum label
	\item Time-synchronous beam-search.
	\item Time-synchronous beam-search with lexical prefix tree and optional 4-gram \gls{lm}.
\end{enumerate}
Phoneme labels are only supported when using prefix tree based decoding.

\subsection{Full-Context RNN-T}

The standard \gls{rnnt} system is built analogously to the \gls{ctc} system. 
The prediction network consists of a single LSTM layer with a hidden dimension of 512. 
The joiner network is a single linear mapping with a ReLU activation function and an output dimension of 640, yielding a total parameter count of 80M.
We use two different decoding modes for the \gls{bpe} version:
\begin{enumerate}[leftmargin=2em]
	\item Time-synchronous beam-search.
	\item Time-synchronous beam-search with lexical prefix tree and optional 4-gram \gls{lm} including zero-encoder internal language model correction~\cite{variani2020hybrid,meng2021internal}.
\end{enumerate}

\subsection{Context-1 Monotonic RNN-T}

The monotonic transducer \cite{tripathi2019monotonic} uses a similar encoder to the \gls{ctc} and \gls{rnnt} models.
It has a context limited to just one history label which is embedded with an embedding size of 256 and then forwarded through a feed-forward prediction network with 2 layers of size 640 and tanh activation.
The joiner network consists of a single layer of size 1024, also with tanh activation.
In total, this leads to a parameter count of around 79M.
In addition to \gls{bpe} labels the context-1 transducer also supports phoneme labels.
The decoding options are identical to the full-context transducer.

\subsection{Attention Encoder-Decoder}

Our \gls{aed} system uses the same features and encoder as our \gls{ctc} system. 
For the decoder, we use a 6-layer Transformer with a dimension of 512.
The model has 101M parameters and uses the same number of layers and 
hidden sizes as the 100M parameter model from the original Loquacious publication.
The model is trained using cross-entropy loss with label smoothing and two CTC auxiliary losses on the 4th and 8th encoder layer.
Decoding is done without external \gls{lm} using a custom implementation of label-synchronous beam search.
This is different from the original Loquacious publication, which used joint \gls{aed}/\gls{ctc} decoding.

\subsection{Factored Hybrid}
Our diphone \gls{fh} has an overall number of 75M parameters.
Decoding is performed using time-synchronous beam search based on dynamic programming and lexical prefix trees.
It is important to note that consistent use of pronunciation variants is only well-defined for this model, as it does not include a blank label.
This is due to the fact that the context dependency modeled on the static decoder structure, as done in the classic weighted finite state framework, is problematic when a blank label is introduced.
In \Cref{subsec:pronvar}, we show the effect of adding pronunciation variants in decoding only.

\section{Experimental Results}

\subsection{Baselines}

We first evaluate \gls{ctc}, \gls{rnnt} and \gls{aed} models without the introduced lexicon and \glspl{lm} for \textit{train.small} and \textit{train.medium}.
\Cref{tab:baseline} shows the results for each of the architectures with \gls{bpe} labels and with greedy decoding or optimal beam search.
Here, we do not include results of the original Loquacious publication, as they did not use any data augmentation, which is essential for good results on smaller datasets.
See \Cref{sec:regularization} for experiments on the effect of regularization techniques and a fair comparison to the original publication.
We tested 128, 256, 512, and 1k labels for \gls{ctc} and \gls{rnnt} and 1k, 2k, and 10k for \gls{aed}.
For CTC and RNN-T, 128 was always slightly better than larger sizes, while for AED 1k is optimal for \textit{train.small} and 10k for \textit{train.medium}.
From the comparison of different \gls{asr} systems on the different test sets, it is clearly visible that it is important to have a task with various testing conditions and different training data scales.
For systems trained on \textit{train.medium}, the results on LibriSpeech and VoxPopuli are very homogeneous for AED and RNN-T.
For Yodas and Commonvoice, there are larger differences.
When switching from \textit{small} to \textit{medium}, doing beam search became more important for both RNN-T and AED on Yodas, but less relevant on LibriSpeech and VoxPopuli.

\subsection{Count-based Language Model}

\begin{table*}[t]
	\begin{center}
		\caption{
			Recognition results for different \gls{bpe} ASR systems to compare the effect of vocabulary restricted search and the addition of the pruned 4-gram \gls{lm}. Abbreviations for test sets: Commonvoice (CV), LibriSpeech (LS), VoxPopuli (VP), Yodas (YD).
			}
		\vspace{0.5em}
		\label{tab:lmvocab}
		\begin{tabular}{|c|c|c|c||c|c||c|c|c|c|}
			\hline
			\multirow{3}{*}{\makecell{Data}} & \multirow{3}{*}{\makecell{Architecture}} & \multirow{3}{*}{\makecell{Vocab}} & \multirow{3}{*}{\makecell{LM}} & \multicolumn{6}{c|}{WER [\%]}\T\\ \cline{5-10}
			& & & & \multicolumn{2}{c||}{Loquacious} & CV & LS & VP & YD\T\\ \cline{5-10}
			& & & & dev & test & \multicolumn{4}{c|}{test}\T\\
			\hline 
			\hline
			\multirow{9}{*}{\makecell{\textit{train.small}\\250h}} & \multirow{3}{*}{\makecell{CTC}} & open & \multirow{2}{*}{no} &  17.2 & 18.6 & 30.1 & 16.7 & 12.5 & 21.1\T\\
			\cline{3-3}
			\cline{5-10}
			& & \multirow{2}{*}{closed} & &  16.2 & 17.3 & 28.2 & 15.4 & 11.7 & 19.9\T\\
			\cline{4-10}
			& & & yes &  12.6 & 13.8 & 21.8 &11.6 &10.2 & 17.2\T\\
			\cline{2-10}
			& \multirow{3}{*}{\makecell{RNN-T}} & open & \multirow{2}{*}{no} & 14.9 & 15.9	& 26.3 & 13.9 & 10.9 & 17.6\T\\
			\cline{3-3}
			\cline{5-10}
			& & \multirow{2}{*}{closed} & & 14.6 & 15.5 & 25.7 & 13.6 & 10.7 & 17.2\T\\
			\cline{4-10}
			& & & yes &  13.1 & 14.1 &22.9 &11.9 &10.1 &16.5\T\\
			\cline{2-10}
			& \multirow{3}{*}{\makecell{mRNN-T}} & open & \multirow{2}{*}{no} & 15.4 & 16.5 & 27.0& 14.6 & 11.2 & 18.6\T\\
			\cline{3-3}
			\cline{5-10}
			& & \multirow{2}{*}{closed} & & 14.7	 & 15.7 & 25.6 & 13.8 & 10.8 & 18.2\T\\
			\cline{4-10}
			& & & yes & 12.2 & 13.4	& 21.5 & 11.1 & \phantom{0}9.7 &16.7\T\\
			\hline
			\hline
			\multirow{9}{*}{\makecell{\textit{train.medium}\\2500h}} & \multirow{3}{*}{CTC} & open & \multirow{2}{*}{no} & 10.6 & 11.5 & 19.3 & \phantom{0}9.0 & \phantom{0}8.1 & 15.4\T\\
			\cline{3-3}
			\cline{5-10}
			& & \multirow{2}{*}{closed} & &  10.0 & 10.8 &18.2 & \phantom{0}8.4 &\phantom{0}7.7 &14.4\T\\
			\cline{4-10}
			& & & yes &  \phantom{0}8.4 & \phantom{0}9.2 & 14.6 & \phantom{0}6.8 & \phantom{0}7.4 & 12.4\T\\
			\cline{2-10}
			& \multirow{3}{*}{\makecell{RNN-T}} & open & \multirow{2}{*}{no} & \phantom{0}8.5 & \phantom{0}9.3 & 15.4 & \phantom{0}7.1 & \phantom{0}7.0 &12.6\T\\
			\cline{3-3}
			\cline{5-10}
			& & \multirow{2}{*}{closed} & & \phantom{0}8.4 & \phantom{0}9.2 & 15.3 &\phantom{0}7.0 & \phantom{0}6.9 &12.5\T\\
			\cline{4-10}
			& & & yes &\phantom{0}8.0	 & \phantom{0}8.8 &14.0 &\phantom{0}6.4 &\phantom{0}6.8 &14.0\T\\
			\cline{2-10}
			& \multirow{3}{*}{\makecell{mRNN-T}} & open & \multirow{2}{*}{no} & \phantom{0}9.1 & 10.2 & 16.3 & \phantom{0}7.3 & \phantom{0}7.3 & 18.8\T\\
			\cline{3-3}
			\cline{5-10}
            & & \multirow{2}{*}{closed} & & \phantom{0}8.9 & 10.0 & 15.8 & \phantom{0}7.1 & \phantom{0}7.2 & 18.7\T\\
			\cline{4-10}
			& & & yes &\phantom{0}8.0 & \phantom{0}8.8 & 14.3 & \phantom{0}6.4 & \phantom{0}7.1 & 12.0\T\\
			\hline
		\end{tabular}
        \vspace{-1em}
	\end{center}
\end{table*}
We use the 216k words vocabulary from \Cref{sec:vocab} and the pruned 4-gram \gls{lm} from \Cref{sec:countlm} to further improve on the baseline results as shown in \Cref{tab:lmvocab}.
As expected using a lexical tree-based search with a language model yields much better results for models trained on \textit{train.small}. 
Surprisingly, just performing vocabulary restricted search without a language model already improved the \gls{bpe}-based model as well.
We see such improvements for all systems trained on \textit{train.small} as well as for \gls{ctc} on \textit{train.medium}. 
But even the context-dependent transducer models still showed slight improvements on the \textit{train.medium}.
When adding the 4-gram \gls{lm}, all models largely improve.
We also tested for a shift in results when we optimize the LM-scale and prior scale for a \gls{ctc} not on the combined dev set, but on each of the subsets.
For  \gls{ctc} trained on \textit{train.small}, we observed that different scales yield an improvement from 17.2 to 16.6 on the Yodas test set.

\Cref{tab:lmmodels} shows \gls{bpe}-\gls{ctc} \gls{asr} results using the four \glspl{lm}, for which the statistics were presented in \Cref{tab:lmppl}.
Despite the difference in \gls{lm} perplexities, we de do not see a substantial difference in the total Loquacious WER, but only for the Yodas test set.
When adding the official LibriSpeech LM data\footnote{\url{https://www.openslr.org/11/}},  we only see improvements on the LibriSpeech test set.
The other sets remain the same, or in case of Yodas, even slightly degrade.
Given that the \gls{lm} with LibriSpeech data is 20 times larger (6GB instead of 300MB) than the 3-gram, it does not really provide much additional benefit.

\begin{table}
	\begin{center}
		\caption{Recognition results for different $n$-gram \glspl{lm} with a \gls{bpe}-\gls{ctc} on \textit{train.medium}.
		}
		\vspace{0.5em}
		\label{tab:lmmodels}
		\begin{tabular}{|c||c|c||c|c|}
			\hline
			\multirow{3}{*}{LM} &  \multicolumn{4}{c|}{WER[\%]} \T\\ \cline{2-5}
			  & \multicolumn{2}{c||}{Loquacious} & LS & YD\T\\
			\cline{2-5}
			& dev & test & \multicolumn{2}{c|}{test}\T\\
			\hline 
			\hline
			3-gram pruned & 8.4 & 9.2 & 6.8 & 12.8\T\\
			\hline
			4-gram pruned & 8.4 & 9.2 & 6.8 & 12.4\T\\
			\hline
			 +  LibriSpeech & 8.3 & 9.1 & 6.5 & 12.9\T\\
			 \hline
			4-gram unpruned & 8.3 & 9.1 & 6.8 & 12.0\T\\
			\hline
		\end{tabular}
		\vspace{-1em}
	\end{center}
\end{table}

%

\subsection{Phoneme-Lexicon}

\begin{table*}[t]
	\begin{center}
		\caption{Recognition results for different \gls{asr} systems comparing \gls{bpe} to phoneme performance. Results of \gls{fh} are excluding speed perturbation. All results include decoding with the pruned 4-gram \gls{lm}.  Abbreviations for test sets: Commonvoice (CV), LibriSpeech (LS), VoxPopuli (VP), Yodas (YD).
		}
		\vspace{0.5em}
		\label{tab:lmlexicon}
		\begin{tabular}{|c|c|c||c|c||c|c|c|c|}
			\hline
			\multirow{3}{*}{\makecell{Data}} &
			\multirow{3}{*}{\makecell{Architecture}} &
			\multirow{3}{*}{\makecell{Label}} &
			\multicolumn{6}{c|}{WER [\%]}\T\\
			\cline{4-9}
			& & & \multicolumn{2}{c||}{Loquacious} & CV & LS & VP & YD\T\\
			\cline{4-9}
			& & & dev & test & \multicolumn{4}{c|}{test}\T\\
			
			\hline 
			\hline
			\multirow{5}{*}{\makecell{\textit{train.small}\\250h}} & \multirow{2}{*}{\makecell{CTC}} & BPE & 12.6 & 13.8 & 21.8 &11.6 &10.2 & 17.2\T\\
			\cline{3-9}
			& & Phon & 12.3 & 13.4 & 21.0 & 11.2 & 10.3 & 15.9\T\\
			\cline{2-9}
			& \multirow{2}{*}{\makecell{mRNN-T}} & BPE & 12.2 & 13.4 & 21.5 & 11.1 & \phantom{0}9.7 &16.7\T\\
			\cline{3-9}
			& & Phon & 12.5	& 13.6 & 21.6 &11.2 &10.4 &16.6\T\\
			\cline{2-9}
			& FH & Phon & 12.5& 13.5 & 22.0& 11.2& 10.2& 15.4\T\\
			\hline
			\hline
			\multirow{5}{*}{\makecell{\textit{train.medium}\\2500h}} & \multirow{2}{*}{CTC} & BPE & \phantom{0}8.4 & \phantom{0}9.2 & 14.6 & \phantom{0}6.8 & \phantom{0}7.4 & 12.4\T\\
			\cline{3-9}
			& & Phon & \phantom{0}8.8 & \phantom{0}9.8 & 15.2 & \phantom{0}7.4 & \phantom{0}7.8 & 13.7\T\\
			\cline{2-9}
			& \multirow{2}{*}{\makecell{mRNN-T}} & BPE & \phantom{0}8.0 & \phantom{0}8.8 & 14.3 & \phantom{0}6.4 & \phantom{0}7.1 & 12.0\T\\
            \cline{3-9}
			& & Phon & \phantom{0}8.5 & \phantom{0}9.2 & 14.5 &\phantom{0}6.8 & \phantom{0}7.7 & 12.1\T\\
			\cline{2-9}
			& FH & Phon &\phantom{0}9.0 &\phantom{0}9.7 & 15.2&\phantom{0}7.4 &\phantom{0}7.9 &13.0\T\\
			\hline
		\end{tabular}
		\vspace{-1em}
	\end{center}
\end{table*}

Using the \gls{ctc} system, we compare the performance when replacing \gls{bpe} labels with phonemes using the created pronunciation lexicon from \Cref{sec:lex}.
\Cref{tab:lmlexicon} shows the results of the phoneme-based \gls{ctc} systems compared to the BPE labels.
It can be seen that when training on \textit{train.small} only the phonetic-based \gls{ctc} system slightly outperforms its \gls{bpe}-based counterpart.
On \textit{train.medium}, the phonetic-based system variant is slightly worse compared to its \gls{bpe} counterpart for all architectures.
Nevertheless, the performance gap is small enough so that the phonetic-based systems can be used for scenarios where this brings advantage, as e.g. for applications where custom pronunciation lexicon entries are desired, or for using systems with more explicit alignment modeling such as \gls{fh}.

\subsection{Restricted Vocabulary}
To verify that it is actually necessary to expand the vocabulary and lexicon beyond what is given in the CMUdict, we perform recognition with a lexicon and LM that was restricted to words in the CMUdict. 
\Cref{tab:cmuonly} shows the results for the \gls{ctc} model trained on \textit{train.small}.
There is a noticeable degradation, showing the justification to expanding the original CMUdict lexicon via \gls{g2p}.

%

\begin{table}
	\begin{center}
		\caption{Recognition results and OOV rates with the BPE-CTC using a restricted lexicon on \textit{train.small}.}
		\label{tab:cmuonly}
		\setlength{\tabcolsep}{0.4em}
		\begin{tabular}{|c||c|c||c|c|}
			\hline
			\multirow{3}{*}{Test-Set} & \multicolumn{4}{c|}{Lexicon}\T\\
			\cline{2-5}
			& \multicolumn{2}{c||}{CMUdict} & \multicolumn{2}{c|}{Full}\T\\
			\cline{2-5}
			&OOV$_{[\%]}$& WER$_{[\%]}$ & OOV$_{[\%]}$ & WER$_{[\%]}$ \T\\
			\hline 
			\hline
			Loq. & 1.85 & 14.5 & 0.47 & 13.8 \T\\
			\hline
			CV & 2.53 & 22.9 & 0.99 & 21.8 \T\\
			\hline
			LS & 2.00 & 12.4 & 0.50 & 11.6 \T\\
			\hline
			VP & 1.33 & 10.6 & 0.11 & 10.2 \T\\
			\hline
			YD & 1.46 & 17.3 & 0.29 & 17.2 \T\\
			\hline
		\end{tabular}
		\vspace{-1em}
	\end{center}
\end{table}

\subsection{Pronunciation Variants}
\label{subsec:pronvar}

The \gls{ctc} pipeline does not support pronunciation variants, so we use the \gls{fh} pipeline to check if pronunciation variants generated by the G2P process are relevant or not.
The lexica are defined as described in \Cref{sec:lex}, so one without and two with additional G2P generated pronunciation variants.
We train two models, one without seeing any pronunciation variants, and one without variants for words that are not in CMUdict.
The recognition process either includes the lexicon that was used for training, or the augmented lexica with different probability masses.
\Cref{tab:pronvar} shows that adding pronunciations during decoding helps, or does not lead to significant degradation.
The ratio of words with G2P generated pronunciations in \textit{dev.all} is only 0.6\%, and only about 50\% of those get a second pronunciation, so the potential of improving accuracy by adding more variants is rather limited.
Moreover, the model trained on single pronunciations generally perform better.
We believe this might be due to the additional alignment paths corresponding to the variants, which might affect the convergence of the model trained from scratch.

\begin{table}
	\begin{center}
		\setlength{\tabcolsep}{0.3em}\renewcommand{\arraystretch}{0.8} 
		\caption{Absolute errors for two \gls{fh} models trained on \textit{train.small} using different lexica. We show the effect of addition of pronunciation variants during recognition.} 
		\vspace{0.5em}
		\label{tab:pronvar}
		\begin{tabular}{|c|c||c|c|c|c|}
			\hline
			\multicolumn{2}{|c||}{Pronunciation Variant} & \multirow{2}{*}{CV} & \multirow{2}{*}{LS} & \multirow{2}{*}{VP} & \multirow{2}{*}{YD}\T\\
			\cline{1-2}
			\makecell{Train} & \makecell{Recog} & & & & \T\\
			\hline
			\hline
			\multirow{3}{*}{\makecell{None}} & None & 5257 & 5449 & 4340 & 867\T\\
			\cline{2-6}
			& 60\% mass & 5234 & 5439 & 4347 & 868\T\\
			\cline{2-6}
			& 80\% mass & 5190 & 5426 & 4336  & 898\T\\
			\hline \hline
			\multirow{3}{*}{\makecell{CMUdict}} & CMUdict& 5325 & 5567 & 4403 & 882\T\\
			\cline{2-6}
			&  60\% mass  & 5324 & 5557 & 4403 & 875\T\\
			\cline{2-6}
			&  80\% mass  & 5305 & 5551 & 4391 & 871\T\\
			\hline
		\end{tabular}
		\vspace{-1em}
	\end{center}
\end{table}


\subsection{Regularization and Overfitting Behavior}
\label{sec:regularization}

The original Loquacious publication reported only results without data augmentation, with the exception of their largest system trained on \textit{train.large}.
We tested the effect of SpecAugment and speed perturbation, which are standard augmentations that were used for past LibriSpeech, TED-Lium and Switchboard systems. 
\Cref{tab:regularizion} shows the results for the \gls{aed} and for the \gls{ctc} system with 4-gram LM when disabling SpecAugment and speed perturbation.
It is widely assumed that data augmentation is essential when working with rather small data sizes of a few hundred or thousand hours.
We also observed substantial improvements for \textit{train.medium}.

	\begin{table}
			\begin{center}
				\caption{Recognition results in WER (\%) on the \textit{train.small} subset for removing SpecAugment and speed perturbation. \textit{Original} refers to the 100M system in the original Loquacious publication.}
				\vspace{0.5em}
				\label{tab:regularizion}
					\begin{tabular}{|c|c|c||c|c|}
							\hline
							\multirow{2}{*}{Architecture} & \multirow{2}{*}{\makecell{Spec-\\Aug.}} &\multirow{2}{*}{\makecell{Speed\\Pert.}}  & \multicolumn{2}{c|}{Loquacious}\T\\
							\cline{4-5}
								& & & dev & test \T\\
							\hline 
							\hline
							\textit{Original} & \multirow{2}{*}{\makecell{no}} &  \multirow{3}{*}{\makecell{no}} & 22.3 & 23.8 \T\\
							\cline{1-1}
							\cline{4-5}
							\multirow{3}{*}{\makecell{AED}} & & & 21.4 & 23.1\T\\
							\cline{2-2}
							\cline{4-5}
							& \multirow{2}{*}{yes} & & 15.1 & 16.6\T\\
							\cline{3-5}
							& & yes & 14.7 & 16.0\T\\
							\hline				
							\hline
							\multirow{3}{*}{CTC + LM} & no &  \multirow{2}{*}{no} & 19.1 & 20.6 \T\\
							\cline{2-2}
							\cline{4-5}
							& \multirow{2}{*}{yes} & & 12.9 & 13.9 \T\\
							\cline{3-5}
							& & yes & 12.6 & 13.8 \T\\
							\hline
						\end{tabular}
						\vspace{-1em}
				\end{center}
			
		\end{table}


\section{Further Analysis and Statistics}

In this section, we want to share some of the insights we gained while working with Loquacious, which are meant to show that it presents
interesting challenges which are not present in more artificial datasets like LibriSpeech.

\subsection{Detailed Word Error Rates}
\label{subsec:detailed-wer}


\begin{table*}[t]
	\begin{center}
		\caption{
			Detailed WERs in terms of substitutions, insertions, and deletions on the LibriSpeech and Yodas dev sets for different models trained on \textit{train.small}. 
			CTC, mRNN-T, and FH results are with 4-gram LM. In case of phonemes, we use end-of-word augmentation.
		}·
		\vspace{0.5em}
		\label{tab:detailed-wer-small}
		\begin{tabular}{|c|c|c||c|c|c|c|c|c|c|c|}
			\hline
			\multirow{3}{*}{\makecell{Architecture}} & \multicolumn{2}{c||}{\multirow{2}{*}{Labels}} & \multicolumn{8}{c|}{WER [\%]}\T\\
			\cline{4-11}
			& \multicolumn{2}{c||}{} & \multicolumn{4}{c|}{LibriSpeech} & \multicolumn{4}{c|}{Yodas}\T\\
			\cline{2-11}
			& Tokens & Number & Sub & Del & Ins & $\Sigma$ & Sub & Del & Ins & $\Sigma$\T\\
			\hline
			\hline
			AED & \multirow{2}{*}{BPE} & 1K & 10.9 & \phantom{0}1.1 & \phantom{0}1.7 & 13.7 & \phantom{0}9.9 & \phantom{0}4.5 & \phantom{0}3.9 & 18.3 \T\\
			\cline{1-1} \cline{3-11}
			\multirow{2}{*}{\makecell{CTC}} & & 128 & \phantom{0}8.9 & \phantom{0}1.3 & \phantom{0}1.1 & 11.3 & \phantom{0}8.3 & \phantom{0}6.4 & \phantom{0}2.1 & 16.8\T\\
			\cline{2-11}
			& Phonemes & 80 & \phantom{0}8.4 & \phantom{0}1.0 & \phantom{0}1.3 & 10.6 & \phantom{0}8.1 & \phantom{0}5.1 & \phantom{0}2.4 & 15.6\T\\
			\hline
			RNN-T & \multirow{2}{*}{\makecell{BPE}} & \multirow{2}{*}{\makecell{128}} & 11.3 & \phantom{0}1.3 & \phantom{0}1.3 & 13.9 & \phantom{0}9.6 & \phantom{0}5.4 & \phantom{0}3.5 & 18.5\T\\
			\cline{1-1} \cline{4-11}
            \multirow{2}{*}{\makecell{mRNN-T}} & & & \phantom{0}8.4 & \phantom{0}1.3 & \phantom{0}1.0 & 10.7 & \phantom{0}7.0 & \phantom{0}6.7 & \phantom{0}2.2 & 15.9\T\\
            \cline{2-11}
			& \multirow{2}{*}{\makecell{Phonemes}} & \multirow{2}{*}{80} & \phantom{0}8.3 & \phantom{0}1.3 & \phantom{0}1.2 & 10.8 & \phantom{0}7.9 & \phantom{0}6.1 & \phantom{0}2.3 & 16.3\T\\
			\cline{1-1} \cline{4-11}
			FH & & & \phantom{0}8.4& \phantom{0}1.6& \phantom{0}1.2& 11.2& \phantom{0}8.2 & \phantom{0}5.1 & \phantom{0}2.5& 15.8\T\\
			\hline
		\end{tabular}
		\vspace{-1em}
	\end{center}
\end{table*}


\begin{table*}[t]
	\begin{center}
		\caption{
			Detailed WERs on the LibriSpeech and Yodas dev sets for different models trained on \textit{train.medium}.
			CTC results are with 4-gram LM. In case of phonemes, we use end-of-word augmentation.
		}
		\vspace{0.5em}
		\label{tab:detailed-wer-medium}
		\begin{tabular}{|c|c|c||c|c|c|c|c|c|c|c|}
			\hline
			\multirow{3}{*}{\makecell{Architecture}} & \multicolumn{2}{c||}{\multirow{2}{*}{Labels}} & \multicolumn{8}{c|}{WER [\%]}\T\\
			\cline{4-11}
			& \multicolumn{2}{c||}{} & \multicolumn{4}{c|}{LibriSpeech} & \multicolumn{4}{c|}{Yodas}\T\\
			\cline{2-11}
			& Tokens & Number & Sub & Del & Ins & $\Sigma$ & Sub & Del & Ins & $\Sigma$\T\\
			\hline
			\hline
			AED & \multirow{2}{*}{BPE} & 10K & 5.6 & 0.5 & 0.8 & 6.9 & 5.4 & 3.0 & 4.8 & 13.2\T\\
			\cline{1-1} \cline{3-11}
			\multirow{2}{*}{\makecell{CTC}} & & 128 & 7.6 & 0.7 & 0.7 & 9.0 & 8.1 & 6.6 & 2.4 & 17.1\T\\
			\cline{2-11}
			& Phonemes & 80 & 5.7 & 0.6 & 0.8 & 7.0 & 5.8 & 6.0 & 2.3 & 14.0\T\\
			\hline
			RNN-T & BPE & 128 & 5.7 & 0.6 & 0.7 & 7.0 & 5.5 & 3.3 & 3.8 & 12.6\T\\
			\hline
		\end{tabular}
		\vspace{-1em}
	\end{center}
\end{table*}

\Cref{tab:detailed-wer-small,tab:detailed-wer-medium} show the detailed WERs for the LibriSpeech and Yodas dev sets 
for different models and label units trained on the \textit{train.small} and \textit{train.medium} sets, respectively.
For \gls{aed}, we report results for different vocabulary sizes because we observed much worse performance with the 10K vocabulary on \textit{train.small}.
For LibriSpeech, we observe that for a given model the amount of deletions and insertions lie in a similar range.
Furthermore, these numbers are also similar across models.
For Yodas however, we observe different error patterns for different configurations. 
For \textit{train.small} (\Cref{tab:detailed-wer-small}), all models have more deletions than insertions.
For \textit{train.medium} (\Cref{tab:detailed-wer-medium}), \gls{aed} and \gls{rnnt} have more insertions than deletions, while \gls{ctc} has more deletions than insertions.

In order to better understand the error patterns, we manually compared the model outputs to the references.
Regarding the insertion behavior of our models, our experience on LibriSpeech shows that they sometimes transcribe single words as multiple words (e.g.``await'' $\rightarrow$ ``a weight'')
or, in case of \gls{aed}, attend to the same audio segment multiple times, resulting in repeated words.
On the Yodas dev set, we additionally observe ``oscillations'' \cite{frieske2024asr-hallucinations}, i.e. cases where most of the model output is correct, but with the addition of a repeating n-gram.
Most frequently, models insert a name at the beginning of an utterance  (e.g. ``would involve...'' $\rightarrow$ ``Matheus Aaron would involve...''), which is related to the name appearing in training transcripts without being present in the audio (see  \Cref{subsec:peculiarities}).
Interestingly, we observe this issue for all models, except for a variant of \gls{fh} for which we used chunked Viterbi training with a fixed target alignment.

Concerning the case of high deletions, we observe that all models omit large sections of some utterances in varying degrees.
This behavior is most extreme in case of the \gls{ctc} models, where almost whole utterances are sometimes omitted.
Interestingly, the sequences where this happens are mostly disjoint between the \gls{aed} model on the one side and the \gls{ctc} and \gls{rnnt} models on the other side.
Upon manual inspection, the affected utterances do not seem to be less intelligible than other utterances.
Furthermore, all models have issues when speakers are spelling words letter by letter, in which case however the \gls{aed} model succeeds more often than the other models.

Lastly, it is interesting to observe that with increased training data the ``weaknesses'' of the \gls{aed} and phoneme \gls{ctc} models, i.e. insertions and deletions,
become more pronounced on Yodas but are halved for LibriSpeech (compare \Cref{tab:detailed-wer-small} and \Cref{tab:detailed-wer-medium}).
For all models, except the BPE \gls{ctc}, the main driver for the reduction in WER on Yodas when switching to \textit{train.medium} is the reduction in substitutions.
For the BPE \gls{ctc}, we even observe a small overall degradation with more training data, for which we don't have an explanation yet.

\subsection{Challenges of the Dataset}
\label{subsec:peculiarities}

As noted in the original Loquacious publication, the dataset contains both clean, read speech, as well as more noisy, spontaneous speech, and everything in between.
Even though the authors employed measures like language identification to filter out erroneous segments, there are remaining aspects that make Loquacious more challenging compared to other academic tasks.
In addition to the training data having transcriptions that do not match the audio in content or language, five of the Commonvoice test set utterances contained no speech even though they had non-empty transcriptions.
We manually pickled 30 examples of \textit{train.small} that a \gls{ctc} model transcribed as empty.
We found that all of them contained speech, but 27 of them did not match the provided transcription at all and 3 turned out to be the Indonesian translations of the English speech.
Lastly, we found that some transcripts in the training sets contain names at the beginning of the utterance that are not spoken.
For example, the name ``Matheus Aaron'' is prepended to 10 utterances in \textit{train.small} and 74 utterances in \textit{train.medium}.
We manually confirmed the absence of this name within the audio of the 10 utterances in \textit{train.small}.
Such aspects make the Loquacious dataset more realistic to work with, while still being well defined and normalized.
%
%
%
%
%

\section{Future Work}

Given the short time since the release of the Loquacious dataset, we focused only on the 250 hours and 2.5k hours subsets.
This is fine in order to investigate the Loquacious dataset as replacement for common smaller academic tasks such as LibriSpeech or TED-Lium.
Still, it is important to expand such studies to the full 25k hours dataset.
For this task, in order to investigate improvements with \gls{lm} integration, much larger text corpora are needed which have to be normalized and standardized to fit the Loquacious task.
Also, the current systems might improve with further optimization and tuning specifically for the Loquacious task.

\section{Conclusion}

In this work, we presented additional resources related to ASR training and evaluation with the Loquacious dataset.
The following resources are published online under a permissive license:
\begin{itemize}[leftmargin=1em]
	\item Vocabulary file containing 216k words
	\item Lexicon files containing 223k (no G2P variants) and 280k (with G2P variants) pronunciations
	\item Sequitur G2P model file
	\item A set of different ARPA count-based LMs trained with KenLM:
	\begin{itemize}[leftmargin=1em]
		\item pruned 3-gram ARPA \gls{lm}
		\item pruned 4-gram ARPA \gls{lm}
		\item un-pruned 4-gram ARPA \gls{lm}
	\end{itemize}
\end{itemize}

The creation of such resources allowed us to conduct experiments with a large variety of different settings related to ASR architectures, decoding methods and label types.
We found that on both the 250 hours \textit{train.small} and 2.5k hours \textit{train.medium} subset the usage of a simple count-based LM substantially improves recognition results.
More surprisingly, even just using a word vocabulary constrained search would yield improvements.
We compared BPE to phoneme representations, showing that phonemes only outperform BPE for a CTC model trained on the 250 hours subset.
Still, the phoneme-based systems had a comparable performance, and for certain research or applications it might be beneficial to also explore such systems.
We discussed some aspects that make Loquacious more interesting for research compared to earlier academic datasets or other alternatives.
Finally, we presented first results with multiple architectures that can be used as reference for reasonable \gls{asr} performances on the \textit{train.small} and \textit{train.medium} sub-tasks.

\section{Limitations}

The biggest limitation of this work is, that it excludes the 25k hours \textit{train.large} subset of the Loquacious dataset.
Working on such data needs a much larger amount of technical preparation, model sizes and training times, which were out of the scope for this work.
In particular, we wanted to look at Loquacious as a replacement for smaller academic tasks, rather covering a variety of conditions instead of aiming for a single best result.
What is excluded regarding the variety are further experiments regarding the model size, as
all models are in a similar range of parameters between roughly 80M to 100M.
While for the BPE-models the BPE sizes are approximately tuned to be optimal, there are some edge cases such as character representations which we did not consider yet.
We were not able to consistently publish all the training recipes, which is planned for a further publication. Nevertheless, the continuous progress of our Loquacious pipelines can be found in our public pipeline repository\footnote{\url{https://github.com/rwth-i6/i6_experiments}}.

\section{Acknowledgements}

This work was partially supported by NeuroSys, which as part of the initiative “Clusters4Future” is funded by the Federal Ministry of Education and Research BMBF (funding IDs 03ZU2106DA and 03ZU2106DD), and by the project RESCALE within the program \textit{AI Lighthouse Projects for the Environment, Climate, Nature and Resources} funded by the Federal Ministry for the Environment, Nature Conservation, Nuclear Safety and Consumer Protection (BMUV), funding ID: 67KI32006A.
The authors gratefully acknowledge the computing time provided to them at the NHR Center NHR4CES at RWTH Aachen University (project number: p0023999). This is funded by the Federal Ministry of Education and Research, and the state governments participating on the basis of the resolutions of the GWK for national high performance computing at universities.
We thank Titouan Parcollet for allowing us to host our resources as part of the official Loquacious release, and thank Mattia Di Gangi, Florian Lux and Moritz Gunz for critical proofreading.

\iffinal

\else
The acknowledgements for this work are redacted for the review version.
This is just a generic placeholder.
\fi

\section{Bibliographical References}\label{sec:reference}

\bibliographystyle{lrec2026-natbib}
\bibliography{mybib}

@string{ICASSP = "Proc.\ IEEE ICASSP"}

@string{ICML = "Proc.\ ICML"}

@string{ASRU = "Proc.\ IEEE ASRU"}

@string{INTERSPEECH = "Proc.\ Interspeech"}

@string{SPEECHCOMMUNICATION = "Speech Communication"}

@string{SPECOM = "Proc.\ SPECOM"}

@string{ACL = "Proc.\ ACL"}

@string{SLT = "Proc.\ IEEE SLT"}

@string{EMNLP = "Proc.\ EMNLP"}

@inproceedings{variani2020hybrid,
	title={Hybrid autoregressive transducer (hat)},
	author={Variani, Ehsan and Rybach, David and Allauzen, Cyril and Riley, Michael},
	booktitle=ICASSP,
	pages={6139--6143},
	year={2020},
	doi={10.1109/ICASSP40776.2020.9053600},  
}

@inproceedings{meng2021internal,
  author={Meng, Zhong and Parthasarathy, Sarangarajan and Sun, Eric and Gaur, Yashesh and Kanda, Naoyuki and Lu, Liang and Chen, Xie and Zhao, Rui and Li, Jinyu and Gong, Yifan},
  booktitle=SLT, 
  title={Internal Language Model Estimation for Domain-Adaptive End-to-End Speech Recognition}, 
  year={2021},
  volume={},
  number={},
  pages={243-250},
  doi={10.1109/SLT48900.2021.9383515}}

@inproceedings{raissi2025analysis,
	title={{Analysis of Domain Shift across ASR Architectures via TTS-Enabled Separation of Target Domain and Acoustic Conditions}},
	author={Raissi, Tina and Rossenbach, Nick and Schl{\"u}ter, Ralf},
	booktitle=ASRU,
	year={2025},
	url = {https://arxiv.org/abs/2508.09868},
}

@article{mohri2002weighted,
	title={Weighted finite-state transducers in speech recognition},
	author={Mohri, Mehryar and Pereira, Fernando and Riley, Michael},
	journal={Computer Speech \& Language},
	volume={16},
	number={1},
	pages={69--88},
	year={2002},
	publisher={Elsevier}
}

@inproceedings{jiang2023neural,
	title={A neural time alignment module for end-to-end automatic speech recognition},
	author={Jiang, Dongcheng and Zhang, Chao and Woodland, Philip C},
	booktitle=INTERSPEECH,
	year={2023},
	pages = {1374--1378},
	doi={10.21437/Interspeech.2023-1071},
}

@inproceedings{rousso2024tradition,
  title     = {{Tradition or Innovation: A Comparison of Modern ASR Methods for Forced Alignment}},
  author    = {Rotem Rousso and Eyal Cohen and Joseph Keshet and Eleanor Chodroff},
  year      = {2024},
  booktitle = INTERSPEECH,
  pages     = {1525--1529},
  doi       = {10.21437/Interspeech.2024-429},
}

@inproceedings{bataev25_interspeech,
	title     = {{NGPU-LM: GPU-Accelerated N-Gram Language Model for Context-Biasing in Greedy ASR Decoding}},
	author    = {Vladimir Bataev and Andrei Andrusenko and Lilit Grigoryan and Aleksandr Laptev and Vitaly Lavrukhin and Boris Ginsburg},
	year      = {2025},
	booktitle = INTERSPEECH,
	pages     = {644--648},
	doi       = {10.21437/Interspeech.2025-955},
	issn      = {2958-1796},
}

@inproceedings{heafield-2011-kenlm,
  title = {{K}en{LM}: Faster and Smaller Language Model Queries},
  author = {Heafield, Kenneth},
  booktitle = {Proceedings of the Sixth Workshop on Statistical Machine Translation},
  month = jul,
  year = {2011},
  address = {Edinburgh, Scotland},
  publisher = ACL,
  url = {https://www.aclweb.org/anthology/W11-2123},
  pages = {187--197},
  month_numeric = {7}
}

@misc{kahn2022flashlight,
	title={Flashlight: Enabling Innovation in Tools for Machine Learning},
	author={Jacob Kahn and Vineel Pratap and Tatiana Likhomanenko and Qiantong Xu and Awni Hannun and Jeff Cai and Paden Tomasello and Ann Lee and Edouard Grave and Gilad Avidov and Benoit Steiner and Vitaliy Liptchinsky and Gabriel Synnaeve and Ronan Collobert},
	year={2022},
	howpublished={Preprint arXiv: 2201.12465},
	url = {https://www.arxiv.org/abs/2201.12465},
}

@INPROCEEDINGS{chan2016las,
  author={Chan, William and Jaitly, Navdeep and Le, Quoc and Vinyals, Oriol},
  booktitle=ICASSP, 
  title={Listen, attend and spell: A neural network for large vocabulary conversational speech recognition}, 
  year={2016},
  volume={},
  number={},
  pages={4960-4964},
  doi={10.1109/ICASSP.2016.7472621}}

@article{Bisani-2008-Joint-sequencemodel,
  title     = {Joint-sequence models for grapheme-to-phoneme conversion},
  author    = {Bisani, Maximilian and Ney, Hermann},
  journal   = SPEECHCOMMUNICATION,
  volume    = {50},
  number    = {5},
  pages     = {434--451},
  year      = {2008},
  month     = {May},
  publisher = {Elsevier {BV}},
  url       = {https://doi.org/10.1016%2Fj.specom.2008.01.002},
  direct    = {S0167639308000046},
  doi       = {10.1016/j.specom.2008.01.002}
}

@inproceedings{parcollet25_interspeech,
  title     = {{Loquacious Set: 25,000 Hours of Transcribed and Diverse English Speech Recognition Data for Research and Commercial Use}},
  author    = {Titouan Parcollet and Yuan Tseng and Shucong Zhang and Rogier C. {van Dalen}},
  year      = {2025},
  booktitle =INTERSPEECH,
  pages     = {4053--4057},
  doi       = {10.21437/Interspeech.2025-720},
  issn      = {2958-1796},
}

@INPROCEEDINGS{librispeech,
  author={Panayotov, Vassil and Chen, Guoguo and Povey, Daniel and Khudanpur, Sanjeev},
  booktitle=ICASSP, 
  title={Librispeech: An ASR corpus based on public domain audio books}, 
  year={2015},
  pages={5206-5210},
  doi={10.1109/ICASSP.2015.7178964}
}

@inproceedings{DBLP:conf/specom/HernandezNGTE18,
  author       = {Fran{\c{c}}ois Hernandez and
                  Vincent Nguyen and
                  Sahar Ghannay and
                  Natalia A. Tomashenko and
                  Yannick Est{\`{e}}ve},
  editor       = {Alexey Karpov and
                  Oliver Jokisch and
                  Rodmonga Potapova},
  title        = {{TED-LIUM} 3: Twice as Much Data and Corpus Repartition for Experiments
                  on Speaker Adaptation},
  booktitle    = {Speech and Computer - 20th International Conference, {SPECOM} 2018,
                  Leipzig, Germany, September 18-22, 2018, Proceedings},
  series       = {Lecture Notes in Computer Science},
  volume       = {11096},
  pages        = {198--208},
  publisher    = {Springer},
  year         = {2018},
  url          = {https://doi.org/10.1007/978-3-319-99579-3\_21},
  doi          = {10.1007/978-3-319-99579-3\_21},
  bibsource    = {dblp computer science bibliography, https://dblp.org}
}

@INPROCEEDINGS{switchboard,
  author={Godfrey, J.J. and Holliman, E.C. and McDaniel, J.},
  booktitle=ICASSP, 
  title={SWITCHBOARD: telephone speech corpus for research and development}, 
  year={1992},
  volume={1},
  number={},
  pages={517-520 vol.1},
  doi={10.1109/ICASSP.1992.225858}
}

@inproceedings{10.1145/1143844.1143891,
	author = {Graves, Alex and Fern\'{a}ndez, Santiago and Gomez, Faustino and Schmidhuber, J\"{u}rgen},
author = {Graves, Alex and Fern\'{a}ndez, Santiago and Gomez, Faustino and Schmidhuber, J\"{u}rgen},
title = {Connectionist temporal classification: labelling unsegmented sequence data with recurrent neural networks},
year = {2006},
isbn = {1595933832},
address = {New York, NY, USA},
url = {https://doi.org/10.1145/1143844.1143891},
doi = {10.1145/1143844.1143891},
pages = {369–376},
numpages = {8},
location = {Pittsburgh, Pennsylvania, USA},
series = {ICML '06}
}

@inproceedings{DBLP:journals/corr/abs-1211-3711,
  author    = {Alex Graves},
  title     = {Sequence Transduction with Recurrent Neural Networks},
  year      = {2012},
  booktitle = {RL 2012 - ICML Workshop on Representation Learning},
  url = {https://www.cs.toronto.edu/~graves/icml_2012.pdf},
  location = {Edinburgh, United Kingdom}
}

@inproceedings{ko15_interspeech,
  title     = {Audio augmentation for speech recognition},
  author    = {Tom Ko and Vijayaditya Peddinti and Daniel Povey and Sanjeev Khudanpur},
  year      = {2015},
  booktitle = INTERSPEECH,
  pages     = {3586--3589},
  doi       = {10.21437/Interspeech.2015-711},
  issn      = {2958-1796},
}

@misc{DBLP:journals/corr/abs-1904-08779,
  author    = {Daniel S. Park and
               William Chan and
               Yu Zhang and
               Chung{-}Cheng Chiu and
               Barret Zoph and
               Ekin D. Cubuk and
               Quoc V. Le},
  title     = {SpecAugment: {A} Simple Data Augmentation Method for Automatic Speech
               Recognition},
  howpublished = {Preprint arXiv: 1904.08779},
  year      = {2019},
  url={https://arxiv.org/abs/1904.08779}, 
}

@inproceedings{raissi2020fh,
  title     = {Context-Dependent Acoustic Modeling Without Explicit Phone Clustering},
  author    = {Tina Raissi and Eugen Beck and Ralf Schlüter and Hermann Ney},
  year      = {2020},
  booktitle = INTERSPEECH,
  pages     = {4377--4381},
  doi       = {10.21437/Interspeech.2020-1244},
}

@book{Bourlard+Morgan:1993,
	title        = {{Connectionist Speech Recognition: a Hybrid Approach}},
	author       = {Bourlard, Herv\'{e} A. and Morgan, Nelson},
	year         = 1993,
	publisher    = {{Kluwer Academic Publishers}},
	address      = {{Norwell, MA}},
	pages        = 352
}

@inproceedings{rybach2011rasr,
	title={{RASR}-the {RWTH} {A}achen university open source speech recognition toolkit},
	author={Rybach, David and Hahn, Stefan and Lehnen, Patrick and Nolden, David and Sundermeyer, Martin and T{\"u}ske, Zoltan and Wiesler, Simon and Schl{\"u}ter, Ralf and Ney, Hermann},
	booktitle=ASRU,
	year={2011}
}

@inproceedings{zeyer2018:returnn,
	title        = {{RETURNN as a Generic Flexible Neural Toolkit with Application to Translation and Speech Recognition}},
	author       = {Zeyer, Albert and Alkhouli, Tamer and Ney, Hermann},
	year         = 2018,
	month        = jul,
	booktitle    = ACL,
	address      = {{Melbourne, Australia}},
	pages        = {128--133},
	doi = "10.18653/v1/P18-4022",
}

@inproceedings{DBLP:conf/emnlp/PeterBN18,
  author    = {Jan{-}Thorsten Peter and
               Eugen Beck and
               Hermann Ney},
  title     = {Sisyphus, a Workflow Manager Designed for Machine Translation and
               Automatic Speech Recognition},
  booktitle = {Proceedings of the 2018 Conference on Empirical Methods in Natural
               Language Processing, {EMNLP}},
  pages     = {84--89},
  year      = 2018,
  url       = {https://www.aclweb.org/anthology/D18-2015/},
  bibsource = {dblp computer science bibliography, https://dblp.org}
}

@misc{frieske2024asr-hallucinations,
      title={Hallucinations in Neural Automatic Speech Recognition: Identifying Errors and Hallucinatory Models}, 
      author={Rita Frieske and Bertram E. Shi},
      year={2024},
      howpublished = {Preprint arXiv: 2401.01572},
      url={https://arxiv.org/abs/2401.01572}, 
}

@inproceedings{tripathi2019monotonic,
  author={Tripathi, Anshuman and Lu, Han and Sak, Hasim and Soltau, Hagen},
  booktitle=ASRU,
  title={Monotonic Recurrent Neural Network Transducer and Decoding Strategies},
  year={2019},
  volume={},
  number={},
  pages={944-948},
  doi={10.1109/ASRU46091.2019.9003822},
}

@inproceedings{sennrich-etal-2016-neural,
    title = "Neural Machine Translation of Rare Words with Subword Units",
    author = "Sennrich, Rico  and
      Haddow, Barry  and
      Birch, Alexandra",
    booktitle = ACL,
    month = aug,
    year = "2016",
    pages = "1715--1725",
    doi = "10.18653/v1/P16-1162",
}

@inproceedings{kudo-2018-subword,
    title = "Subword Regularization: Improving Neural Network Translation Models with Multiple Subword Candidates",
    author = "Kudo, Taku",
    editor = "Gurevych, Iryna  and
      Miyao, Yusuke",
    booktitle = ACL,
    month = jul,
    year = "2018",
    address = "Melbourne, Australia",
    url = "https://aclanthology.org/P18-1007/",
    doi = "10.18653/v1/P18-1007",
    pages = "66--75"
}

@inproceedings{conformer,
  author       = {Anmol Gulati and
                  James Qin and
                  Chung{-}Cheng Chiu and
                  Niki Parmar and
                  Yu Zhang and
                  Jiahui Yu and
                  Wei Han and
                  Shibo Wang and
                  Zhengdong Zhang and
                  Yonghui Wu and
                  Ruoming Pang},
  title        = {Conformer: Convolution-augmented Transformer for Speech Recognition},
  booktitle    = INTERSPEECH,
  pages        = {5036--5040},
  year         = {2020},
  doi       = {10.21437/Interspeech.2020-3015}
}

\end{document}